\documentclass{bmcart}

\usepackage[utf8]{inputenc} 
\usepackage{caption}

\def\includegraphics{}

\startlocaldefs
\endlocaldefs

\begin{document}

\begin{frontmatter}

\begin{fmbox}
\dochead{Arnekt Solutions}


\title{Deep Learning for Digital Text Analytics : Sentiment Analysis}


\author[
   addressref={aff1},                   
   email={reshma.u@arnekt.com}   
]{\inits{RU}\fnm{} \snm{Reshma U}}
\author[
   addressref={aff1},
   email={barathiganesh.hb@arnekt.com}
]{\inits{HBBG}\fnm{} \snm{Barathi Ganesh H B}}
\author[
   addressref={aff1},
   email={mandar.kale@arnekt.com}
]{\inits{Mandar}\fnm{} \snm{Mandar Kale}}
\author[
   addressref={aff1},
   email={prachi.m@arnekt.com}
]{\inits{Prachi}\fnm{} \snm{Prachi Mankame}}
\author[
   addressref={aff1},
   email={gouri.k@arnekt.com}
]{\inits{Gouri}\fnm{} \snm{Gouri Kulkarni}}


\address[id=aff1]{
  \orgname{Arnekt Solutions Pvt. Ltd.}, 
  \street{Pentagon P-3, Magarpatta City, Pune},                     %
  \city{Maharashtra},                              
  \cny{India}                                    
}



\end{fmbox}


\begin{abstractbox}

\begin{abstract} 
In today's scenario, imagining a world without negativity is something very unrealistic, as bad NEWS spreads more virally than good ones. Though it seems impractical in real life, this could be implemented by building a system using Machine Learning and Natural Language Processing techniques in identifying the news datum with negative shade and filter them by taking only the news with positive shade (good news) to the end user. In this work, around two lakhs datum have been trained and tested using a combination of rule-based and data driven approaches. VADER along with a filtration method has been used as an annotating tool followed by statistical Machine Learning approach that have used Document Term Matrix (representation) and Support Vector Machine (classification). Deep Learning algorithms then came into picture to make this system reliable (Doc2Vec) which finally ended up with Convolutional Neural Network(CNN) that yielded better results than the other experimented modules. It showed up a training accuracy of 96\%, while a test accuracy of (internal and external news datum) above 85\% was obtained.
\end{abstract}


\begin{keyword}
\kwd{Sentiment Analysis}
\kwd{Natural Language Processing}
\kwd{Convolutional Neural Networks}
\kwd{Text Classification}
\kwd{Deep Learning}
\end{keyword}


\end{abstractbox}
%

\end{frontmatter}



\section*{Introduction}
Artificial intelligence(AI) has turned out to be the most desirous field in this information era. Intelligence is nurtured into systems in such a way that it could replicate human behaviour in learning, reasoning, problem solving, perceiving intuitions and also understand human languages. In simple, AI is about machine duplicating human's role not by means of explicitly teaching a system but instead, making use of its sub-discipline Machine Learning(ML). Machine Learning rose up with the need to teach machines from past experiences. ML got rooted strong as things got digitized and became available for data analysis and many more applications. With the emergence of Natural Language Processing(NLP) one of the fields of AI, data in the form of texts seemed to get benefited as NLP relies fully on ML. Deep Learning is the cutting-edge of the cutting-edge field on demand Machine Learning \footnote{https://www.forbes.com/sites/bernardmarr/2016/12/08/what-is-the-difference-between-deep-learning-machine-learning-and-ai/\#4dfe052226cf}. In cases where a thought is required to process an outcome or make a decision, Neural Networks(NN) an imitation of human brain plays the crucial role. Getting to know all the advancements and possibilities in AI, now lets get to know the applications or use cases in one of the sub-fields of AI, Natural Language Processing.

As the name depicts Natural Language Processing (NLP) or Language Processing entirely makes use of human generated text data from various resources, data warehouses and etc., for processing and developing real time applications or even serve as a small means of support for building larger applications. When compared to the amount of data that has been generated on daily basis, text data seems to overreach other datum like image, video and audio. The reason for immense growth of text data is obvious, as we individuals are more drowned with the usage and production of text data than others. The endless growth of text data starts from the very Short-Message-Service(SMS) platform to the data that has been generated via social medias(Whatsapp, Facebook, Twitter, Instagram, LinkedIn and etc,.), blogs, posts, websites, research papers, documents and many more.

\begin{figure}[h!]
  \caption*{ “Data is the new oil. It’s valuable, but if unrefined it cannot really be used. \\ It has to be changed into gas, plastic, chemicals, etc to create a valuable entity\\ that drives profitable activity; so must data be broken down, analyzed for it to \\have value.\footnote{http://www.humbyanddunn.com/}\footnote{https://medium.com/twenty-one-hundred/data-is-the-new-oil-a-ludicrous-proposition-1d91bba4f294}” \csentence{-Clive Humby}}
      \end{figure}


The unstructured data that we generate everyday include emails, text messages, posts, comments and status in social medias, word documents and presentations in business or academic use, mp3 files, images and videos that are shared online and lots more. The generation of such unstructured data seems to be much greater than the structured/semi-structured data such as ATM bills, government applications, admission forms, transaction details and etc,. By knowing the importance of data and in specific, the surplus amount of text data that gains more importance than the other datum (audio, video or image), lets get started with its significance in building real time applications.

\section*{Need for Text Analytics} 
Text based applications in real life has still not gained a state-of-art performance till date. When we compare other datum such as image and speech, the applications built using them such as Face recognition, Speech recognition, Emotion detection from speech signals and some more have attained up to the minute result in its deployment to real time. Where as if we consider a simple Text classification problem which revolves around text data, text ambiguity as a result of holographs, metaphors and many other word forms makes the problem more complicated. System may attain good accuracy with respect to the data we feed to train the system, but this is always not applicable in real world scenario. 

Emotions could be imported from the speech signal by the variation of its amplitude and frequency to some extent. If the same has to be applied for texts, the sarcastic usage of words and ironical word forms makes the emotion detection problem even more trivial. From the above aspects one could clearly identify the importance of working with text data and its complexity in attaining a standard model.

\section*{Text Analytics Applications}
Applications that have been built using text data is once again used on daily basis just like the text data been generated by us on everyday occasions \cite{ganesh2018social}. Predictions of the next word while doing a Google search or sending text messages, spell check and correction of misspelled words while composing an email or document, document classification - to group similar documents together just like creating folders for movies in different languages in ones desktop \cite{reshma2006supervised}, author identification - to identify the author of a document which does not contain the author's name \cite{reshma2015author}, author profiling - finding the age, gender, mother-tongue, depression state and many more personal characteristics of an individual \cite{barathirepresentation}, machine translation - translation from one language to another, information retrieval - document retrieval for legal documents \cite{barathiamrita_cen_nlp}, catchphrase extraction for legal phrase extraction \cite{barathiamrita_cen_nlp}, paraphrase detection, sentiment analysis, question-answering - chat bots, document summarization and plenty of other applications in process and that are yet to come.    

\section*{Sentiment Analysis}
From the above applications our area of focus would be on Sentiment Analysis. As the name depicts this application completely relies in bringing out the emotional state behind the digital text data. Emotions in digital texts could be interpreted by human-beings by understanding the state of mind of the person and the context behind the texts. It becomes rather a simple problem to deal with if dealt between humans. The case is not the same with machines as they barely understand the contexts and emotional state of a person. This field has shown great interest in various domains as our lives have been digitized in all ways from the newspaper we read as a day begins to the materials we purchase and comment on the same. The state where a person liked the product or not could be judged by looking out to the review section in a e-commerce domain (Amazon, Flipkart, and etc,.). In a similar fashion the emotions behind the messages we send our loved ones, the status we update on Social medias (Facebook, Twitter and etc,.), the news we read from news websites and lot more to go. All these define the regular usage of emotions behind the digital text data. These datum are not only useful in detecting the emotions as such but spreads wide to domains of Cyber security, Law and Order, Business statistics, Marketing analysis and many more. In today's scenario this field has turned out to be an important business module as the profit of an organization absolutely deals with the customer satisfaction.

\begin{figure}[h!]
  \caption*{ “This not only provides insight into what people think about your brand, \\but it can go a lot deeper. It can expose why people are thinking it.” \\ \csentence{-Daniel Angus}}
      \end{figure}

Sentiment Analysis has turned to be a tough problem as the structure of the data it deals with in most cases are unstructured \cite{pang2008opinion}. The messages we type, the comments we give on each product or the status we update on Social medias does not have any specific format and language usage as such. People across the world tweet, chat and comment on things with their own languages (mother tongue), or in a global language (English), or a combination of both (mother tongue + global language) or even a native of different origin uses global language. To get a more closer idea about this lets focus on a country like India where there are so many languages with and without explicit scripts. People here use their own languages to express their views (Hindi, Marati, Tamil, Malayalam and etc,.). A native speaker of Hindi for example will probably have an effect of their native language on English language usage. It holds the same for other native language users all around the world. Apart from different language usage, dialects of languages are also around for a long time. People even use their native language along with English language that results in a code-mixed data to convey messages. This makes the structure of text data even more unlike from regular language structure. Aside from these possibilities, the usage of emoticons which has been trending since 1990's gives a different context and ironical touch to text data collected from Social medias in specific. This symbolizes the complexity involved in the structure of the texts and the hidden emotions present in them.

\section*{Problem Formulation}

\begin{figure}[h!]
  \caption*{ “We could covert a news articles about a company into a number \\that expresses the positive sentiment ... and negative sentiment” \csentence{-Liang Zhou}}
      \end{figure}

In the current trending digital age, events which we come across our daily life starts from the sharing of information. Information that we commonly share include current trending NEWS in almost all domains (Politics, Sports, Cinema,etc,.) which sometimes turns off or turns on our day. Apart from updating ourselves with the information that we come across via sharing, there are people who show genuine interest in updating current events by reading news from news sites and news apps. Imagine a world where no crime exists and people share and read only good news all around. This is something that is highly impossible in real life, but wonder if there exists a intelligence where only good news pops up each time you open it! This could be made possible by means of Artificial Intelligence (Natural Language Processing(NLP) and Machine Learning(ML)). By applying Natural Language Processing techniques to find the emotions behind the context of a text and display only the texts with positive outlook will thereby help in executing the events which are impossible to cut down in real life. Here we aim at bringing out only the positive contents from news data, by ignoring the ones which have a negative shade on it to the end user.

News data which are open to all domains (Sports, Cinema, Politics, Technology and etc.) were crawled from Times Of India, BBC News, India Today, IndiaTV, Reuters, The Hindu, STAR News and etc., during the period of July 2017 to January 2018. We also collected news from other news portals like CNN-IBN, BBC and etc., which were found to be a lookalike of the other news which were previously collected and hence we ignored these set of news. The total of 1,84,707 news were collected during the mentioned period.

\subsection*{Corpus Complexity}
Unlike the structure of social media texts which we have seen earlier, the structure of news data is different. Specific parameters could not be tuned to extract information from these as there are no emoticons, hash tags and code-mixed data which are very specific in case of social media texts. Dealing with the of identification of sentiments hidden behind the news data is not an easy task as the sarcasm behind the news data cannot be evaluated by a normal human himself/herself if they are not aware of events happening around them. News which has been crawled between a specific period does not include the dependencies associated with it. The synopsis here is that the classification of positive and negative news depends entirely on a common users point of view and his/her awareness about the happenings. In such a state defining the positive/negative news are difficult, but news which contains negative shade in it could be identified to some extent and sent for reviewing (human intervention) by filtering the ones considered as negative or neutral before passing to the end viewers.   

\subsection*{VADER}
VADER - Valence Aware Dictionary and sEntiment Reasoner can be used as an annotating tool for the lakhs of news data that has been crawled from news websites between the period Jul 2017 – Jan 2018 \cite{gilbert2014vader}. Though this tool has been specifically designed to deal with social media texts, this could be used in annotating news datum as it could describe the news with the negative and positive shade to some extent based on the word count of the same. VADER is a library that follows the lexical approach and has the advantage of self accessing the sentiment of any given text without the need of any previously labelled text data unlike Machine Learning approach. The score based methodology ranges from -4 for most negative word to +4 for the most positive word, while 0 remains neutral. The sum of the scores of these lexical features would eventually result in the sentiment score of a sentence. VADER not only takes lexical features for classifying the data, but also makes use of contextual features like capitalization, punctuation, degree modifiers, shift in polarity due to the word like "but" and examining the tri-gram before a sentiment-laden lexical feature to catch polarity negation \footnote{http://datameetsmedia.com/vader-sentiment-analysis-explained/}\cite{gilbert2014vader}. VADER is used here for annotating the crawled data into three different classes (1:Positive, -1:Negative and 0:Neutral). The next idea would be to concentrate mainly on the news data which has been segregated into positive class, as our aim is to filter out news with any kind of negative or neutral shade in it. This would help in building a reliable system using Deep Learning algorithms.

The VADER algorithms maintains the recall of 100\% for the negative news, i.e it is obvious that news with the more negative terms than the positive terms can be considered as the news reflecting the negativity. Using this hypothesis once the corpus have been tagged using the VADER, only the negative and neutral news (taken as the negative news) are used to train the one class Support Vector Machine \cite{manevitz2001one}. 10-fold 10-cross validation is used to train the hyper parameters of SVM \cite{duan2003evaluation}. On successive model creation the filtration on the positive news are carried out.   
\subsection*{Conventional Machine Learning}
Once the annotation is done, the news data has got classified into two groups. Rather than manually annotating the news data for the purpose of descriptive modelling, VADER has broken down the process in a simple manner. By making use of these datum, the system could be trained to study the news with positive and negative shade \cite{gilbert2014vader}. As a matter of choosing a model that would fit into it the data driven method, Document Term Matrix representation with Support Vector Machine classification was chosen \cite{ganesh2016vector, kp2017vector}. Though it followed the classical approach of representation and classification, this model did perform well by getting trained on news data that were annotated using VADER. The flaw it contained was the amount of time it consumed in getting trained to build a suitable model using 10-fold 10-cross validation. The model creation took several days and hence this approach was dropped out. As a next step Deep Learning methods were taken into consideration rather than the traditional Machine Learning approach. 

\begin{figure}[h!]
  \caption*{ “I think people need to understand that deep learning is making a lot of \\things, behind-the-scenes, much better. Deep learning is already working in \\ Google search, and in image search; it allows you to image search a term like \\'hug'.” \csentence{-Geoffrey Hinton}}
      \end{figure}

As a matter of fact Doc2Vec was the next model of interest according to the representational advantage over other methods \cite{barathiamrita_cen_nlp}. As this model is an extension of word2vec with the advantage of following an unsupervised methodology in generating vectors for the sentences or words considered, this method has been chosen to find the similarity between sentences. Another probable reason for selecting this model is that this approach requires data that has been tagged or annotated as these labels serve as one of the feature for further processing. Since this model creates vectors for each sentence and the corresponding labels the memory usage increases thereby increasing the time consumption. This has once again led to switch to a new efficient model that can handle the computational complexity as well as attain a reliable system with favourable outcome. By keeping hold of these constrains several Deep Learning algorithms as well as Graphics Processing Unit (GPU) computing were studied. After analysis and study the Convolutional Neural Network(CNN) which previously worked well for computer vision problems as has now taken a hold in text analytics has been chosen \cite{kim2014convolutional}. By taking a closer look at the size of the data and the time consumption in the previous methodologies GPU (Tesla K20x) has been taken for processing and building CNN model for sentiment analysis.
 
\subsection*{Deep Learning: Convolutional Neural Network}
Convolutional Neural Network (CNN) also known as ConvNets is a Deep Learning tool that has gained expertise in Computer Vision (CV) applications \cite{srinivas2016taxonomy}. The use of neural networks for NLP applications is attracting huge interest in the research community and they are systematically applied to all NLP tasks \cite{kim2014convolutional}. The fundamental idea of ConvNets is to consider feature extraction and classification as one jointly trained task. The scope of using this methodology in text analytics has proven to be advantageous in various ways \cite{dos2014deep, kim2014convolutional, severyn2015twitter}. This idea has been improved over the years, in particular by using many layers of convolutions and pooling to sequentially extract a hierarchical representation of the input. By means of hyper-parameter tuning and a series of forward and back propagation we could end up with the desired output of our choice. 

CNN functions with two main layers, the convolutional layer and the pooling layer. As the name depicts the convolutional layer acts as the central component in building the framework \cite{zhang2015character}. It is in this layer where multiple convolutions are performed using kernel functions in order to attain stack of feature maps. These stacked feature maps that has been obtained from convolutional layers are then passed to an activation function (relu/tanh) which makes the neural network to be highly non-linear and powerful \cite{zhang2015character}. It is to be noted that the kernel size always remains lesser than the input size (text/image/speech) in order to fit into the input space. The size of the feature map is directly proportional to the size of the kernel used. The size of the feature map could be increased or re-sized to the input size by means of padding. This helps in preserving the size of feature map and in avoiding the narrowing of feature map at every layer in CNN. There is yet another parameter in the convolutional layer which helps in moving across the input data according to the number of steps specified. It is called the stride which helps in the movement of kernel across the input which by default is set to 1. The pooling layer concentrates on the dimensionality of the data. It helps in reducing the dimensionality of the data by shortening the training time and also tackles the over-fitting issues if any. The commonly used pooling technique is Max-pooling. Here the pooling window gets slides over the output of the convolutional layer and extracts the layer with maximum turn outs \cite{zhang2015character}. 

\begin{figure}[h!]
  \caption*{ “The pooling operation used in convolutional neural networks is a big mistake,\\ and the fact that it works so well is a disaster.” \csentence{-Geoffrey Hinton}}
      \end{figure}
      
The output from the pooling layer is changed to a 1D array and then sent to the fully connected layers where the entire process gets trained using several forward and backward propagation with gradient descent as a key ingredient. As an added feature, the dropout parameter is used in the training phase for regularization purpose. This gives an accuracy boost by dropping certain neurons in the network based on the its probability. These dropped out neurons are then made active in the next propagation by dropping down the next set of neurons. This operation is applied only to the input and hidden layers in a neural network architecture as a matter of obtaining the desired outcome from the entire process \cite{zhang2015character}. Among the hyper-parameters tuned, pre-trained word vectors (Glove vectors) have been used for better input representation. 

After the optimization of hyper-parameters, the news data has got trained using the final hyper-parameters such as number of filters=600, filter size = 3, pooling strategy = Softmax, activation = Relu, Dropout = 0.5, epoch = 20, strides = 1, padding = valid along with the previously mentioned glove vectors for better input representation. These datum are then validated on 20\% of news data from the total available news data. For better precision in building a reliable system external news data with manual tagging has been used as a next set of test datum. From the results obtained it has been noted that the internal test accuracy and the external test accuracy of the CNN has managed to cross 85\%. This shows that the built model works well with real time news data. The accuracy could be further improved by using train data that are manually tagged along with VADER as the external news data are manually evaluated for producing an accuracy score. 

\section*{Conclusion}
With an aim of identifying the news data with negative shade and discarding them by sending only the positive news datum to the end user from various crawled news has been implemented with an acceptable performance. VADER along with the one class SVM has reduced the workload of manually labelling the datum that has been fetched. As a next step of making use of a data driven model in evaluating the polarities of these tagged news datum, several approaches from the statistical Machine Learning approach(DTM-SVM) to the Neural Network algorithms (doc2vec) have been utilized. As these methods failed to built an efficient model, CNN has been taken into consideration. A combination of CNN with GPU has yielded out better results as compared to previous methods. Training accuracy of 96\% and testing accuracy of internal and external news datum have crossed 85\% of accuracy. This model could be further improved by fine-tuning the train data by including manually annotated data along with the VADER annotated datum. This could improve the performance of the system while dealing with real time datum.

\bibliographystyle{bmc-mathphys} 
\bibliography{bib}      

\begin{thebibliography}{16}
\ifx \bisbn   \undefined \def \bisbn  #1{ISBN #1}\fi
\ifx \binits  \undefined \def \binits#1{#1}\fi
\ifx \bauthor  \undefined \def \bauthor#1{#1}\fi
\ifx \batitle  \undefined \def \batitle#1{#1}\fi
\ifx \bjtitle  \undefined \def \bjtitle#1{#1}\fi
\ifx \bvolume  \undefined \def \bvolume#1{\textbf{#1}}\fi
\ifx \byear  \undefined \def \byear#1{#1}\fi
\ifx \bissue  \undefined \def \bissue#1{#1}\fi
\ifx \bfpage  \undefined \def \bfpage#1{#1}\fi
\ifx \blpage  \undefined \def \blpage #1{#1}\fi
\ifx \burl  \undefined \def \burl#1{\textsf{#1}}\fi
\ifx \doiurl  \undefined \def \doiurl#1{\textsf{#1}}\fi
\ifx \betal  \undefined \def \betal{\textit{et al.}}\fi
\ifx \binstitute  \undefined \def \binstitute#1{#1}\fi
\ifx \binstitutionaled  \undefined \def \binstitutionaled#1{#1}\fi
\ifx \bctitle  \undefined \def \bctitle#1{#1}\fi
\ifx \beditor  \undefined \def \beditor#1{#1}\fi
\ifx \bpublisher  \undefined \def \bpublisher#1{#1}\fi
\ifx \bbtitle  \undefined \def \bbtitle#1{#1}\fi
\ifx \bedition  \undefined \def \bedition#1{#1}\fi
\ifx \bseriesno  \undefined \def \bseriesno#1{#1}\fi
\ifx \blocation  \undefined \def \blocation#1{#1}\fi
\ifx \bsertitle  \undefined \def \bsertitle#1{#1}\fi
\ifx \bsnm \undefined \def \bsnm#1{#1}\fi
\ifx \bsuffix \undefined \def \bsuffix#1{#1}\fi
\ifx \bparticle \undefined \def \bparticle#1{#1}\fi
\ifx \barticle \undefined \def \barticle#1{#1}\fi
\ifx \bconfdate \undefined \def \bconfdate #1{#1}\fi
\ifx \botherref \undefined \def \botherref #1{#1}\fi
\ifx \url \undefined \def \url#1{\textsf{#1}}\fi
\ifx \bchapter \undefined \def \bchapter#1{#1}\fi
\ifx \bbook \undefined \def \bbook#1{#1}\fi
\ifx \bcomment \undefined \def \bcomment#1{#1}\fi
\ifx \oauthor \undefined \def \oauthor#1{#1}\fi
\ifx \citeauthoryear \undefined \def \citeauthoryear#1{#1}\fi
\ifx \endbibitem  \undefined \def \endbibitem {}\fi
\ifx \bconflocation  \undefined \def \bconflocation#1{#1}\fi
\ifx \arxivurl  \undefined \def \arxivurl#1{\textsf{#1}}\fi
\csname PreBibitemsHook\endcsname

\bibitem{ganesh2018social}
\begin{botherref}
\oauthor{\bsnm{Ganesh}, \binits{H.}}, et al.:
Social media analysis based on semanticity of streaming and batch data.
arXiv preprint arXiv:1801.01102
(2018)
\end{botherref}
\endbibitem

\bibitem{reshma2006supervised}
\begin{botherref}
\oauthor{\bsnm{Reshma}, \binits{U.}},
\oauthor{\bsnm{Barathi~Ganesh}, \binits{H.}},
\oauthor{\bsnm{Anand~Kumar}, \binits{M.}},
\oauthor{\bsnm{Soman}, \binits{K.}}:
Supervised methods for domain classification of tamil documents
(2006)
\end{botherref}
\endbibitem

\bibitem{reshma2015author}
\begin{bchapter}
\bauthor{\bsnm{Reshma}, \binits{U.}}, \betal:
\bctitle{Author identification based on word distribution in word space}.
In: \bbtitle{Advances in Computing, Communications and Informatics (ICACCI),
  2015 International Conference On},
pp. \bfpage{1519}--\blpage{1523}
(\byear{2015}).
\bcomment{IEEE}
\end{bchapter}
\endbibitem

\bibitem{barathirepresentation}
\begin{botherref}
\oauthor{\bsnm{Barathi~Ganesh}, \binits{H.}},
\oauthor{\bsnm{Reshma}, \binits{U.}},
\oauthor{\bsnm{Anand~Kumar}, \binits{M.}},
\oauthor{\bsnm{Soman}, \binits{K.}}:
Representation of target classes for text classification-amrita\_cen\_nlp@
  rusprofiling pan 2017
\end{botherref}
\endbibitem

\bibitem{barathiamrita_cen_nlp}
\begin{botherref}
\oauthor{\bsnm{Barathi~Ganesh}, \binits{H.}},
\oauthor{\bsnm{Reshma}, \binits{U.}},
\oauthor{\bsnm{Anand~Kumar}, \binits{M.}},
\oauthor{\bsnm{Soman}, \binits{K.}}:
Amrita\_cen\_nlp@ irled 2017
\end{botherref}
\endbibitem

\bibitem{pang2008opinion}
\begin{barticle}
\bauthor{\bsnm{Pang}, \binits{B.}},
\bauthor{\bsnm{Lee}, \binits{L.}}, \betal:
\batitle{Opinion mining and sentiment analysis}.
\bjtitle{Foundations and Trends{\textregistered} in Information Retrieval}
\bvolume{2}(\bissue{1--2}),
\bfpage{1}--\blpage{135}
(\byear{2008})
\end{barticle}
\endbibitem

\bibitem{gilbert2014vader}
\begin{bchapter}
\bauthor{\bsnm{Gilbert}, \binits{C.H.E.}}:
\bctitle{Vader: A parsimonious rule-based model for sentiment analysis of
  social media text}.
In: \bbtitle{Eighth International Conference on Weblogs and Social Media
  (ICWSM-14). Available at (20/04/16) Http://comp. Social. Gatech.
  Edu/papers/icwsm14. Vader. Hutto. Pdf}
(\byear{2014})
\end{bchapter}
\endbibitem

\bibitem{manevitz2001one}
\begin{barticle}
\bauthor{\bsnm{Manevitz}, \binits{L.M.}},
\bauthor{\bsnm{Yousef}, \binits{M.}}:
\batitle{One-class svms for document classification}.
\bjtitle{Journal of machine Learning research}
\bvolume{2}(\bissue{Dec}),
\bfpage{139}--\blpage{154}
(\byear{2001})
\end{barticle}
\endbibitem

\bibitem{duan2003evaluation}
\begin{barticle}
\bauthor{\bsnm{Duan}, \binits{K.}},
\bauthor{\bsnm{Keerthi}, \binits{S.S.}},
\bauthor{\bsnm{Poo}, \binits{A.N.}}:
\batitle{Evaluation of simple performance measures for tuning svm
  hyperparameters}.
\bjtitle{Neurocomputing}
\bvolume{51},
\bfpage{41}--\blpage{59}
(\byear{2003})
\end{barticle}
\endbibitem

\bibitem{ganesh2016vector}
\begin{bchapter}
\bauthor{\bsnm{Ganesh}, \binits{H.B.}},
\bauthor{\bsnm{Kumar}, \binits{M.A.}},
\bauthor{\bsnm{Soman}, \binits{K.}}:
\bctitle{From vector space models to vector space models of semantics}.
In: \bbtitle{Forum for Information Retrieval Evaluation},
pp. \bfpage{50}--\blpage{60}
(\byear{2016}).
\bcomment{Springer}
\end{bchapter}
\endbibitem

\bibitem{kp2017vector}
\begin{botherref}
\oauthor{\bsnm{KP}, \binits{S.}}, et al.:
Vector space model as cognitive space for text classification.
arXiv preprint arXiv:1708.06068
(2017)
\end{botherref}
\endbibitem

\bibitem{kim2014convolutional}
\begin{botherref}
\oauthor{\bsnm{Kim}, \binits{Y.}}:
Convolutional neural networks for sentence classification.
arXiv preprint arXiv:1408.5882
(2014)
\end{botherref}
\endbibitem

\bibitem{srinivas2016taxonomy}
\begin{barticle}
\bauthor{\bsnm{Srinivas}, \binits{S.}},
\bauthor{\bsnm{Sarvadevabhatla}, \binits{R.K.}},
\bauthor{\bsnm{Mopuri}, \binits{K.R.}},
\bauthor{\bsnm{Prabhu}, \binits{N.}},
\bauthor{\bsnm{Kruthiventi}, \binits{S.S.}},
\bauthor{\bsnm{Babu}, \binits{R.V.}}:
\batitle{A taxonomy of deep convolutional neural nets for computer vision}.
\bjtitle{Frontiers in Robotics and AI}
\bvolume{2},
\bfpage{36}
(\byear{2016})
\end{barticle}
\endbibitem

\bibitem{dos2014deep}
\begin{bchapter}
\bauthor{\bparticle{dos} \bsnm{Santos}, \binits{C.}},
\bauthor{\bsnm{Gatti}, \binits{M.}}:
\bctitle{Deep convolutional neural networks for sentiment analysis of short
  texts}.
In: \bbtitle{Proceedings of COLING 2014, the 25th International Conference on
  Computational Linguistics: Technical Papers},
pp. \bfpage{69}--\blpage{78}
(\byear{2014})
\end{bchapter}
\endbibitem

\bibitem{severyn2015twitter}
\begin{bchapter}
\bauthor{\bsnm{Severyn}, \binits{A.}},
\bauthor{\bsnm{Moschitti}, \binits{A.}}:
\bctitle{Twitter sentiment analysis with deep convolutional neural networks}.
In: \bbtitle{Proceedings of the 38th International ACM SIGIR Conference on
  Research and Development in Information Retrieval},
pp. \bfpage{959}--\blpage{962}
(\byear{2015}).
\bcomment{ACM}
\end{bchapter}
\endbibitem

\bibitem{zhang2015character}
\begin{bchapter}
\bauthor{\bsnm{Zhang}, \binits{X.}},
\bauthor{\bsnm{Zhao}, \binits{J.}},
\bauthor{\bsnm{LeCun}, \binits{Y.}}:
\bctitle{Character-level convolutional networks for text classification}.
In: \bbtitle{Advances in Neural Information Processing Systems},
pp. \bfpage{649}--\blpage{657}
(\byear{2015})
\end{bchapter}
\endbibitem

\end{thebibliography}

\end{document}